# GALGO: A Genetic ALGOrithm Decision Support Tool for Complex Uncertain Systems Modeled with Bayesian Belief Networks

Carlos Rojas-Guzmán
Chemical Engineering Department
Massachusetts Institute of Technology
Cambridge, MA, 02139

Mark A. Kramer
Chemical Engineering Department
Massachusetts Institute of Technology
Cambridge, MA, 02139

## Abstract

Bayesian belief networks can be used to represent and to reason about complex systems with uncertain, incomplete and conflicting information. Belief networks are graphs encoding and quantifying probabilistic dependence and conditional independence among variables. One type of reasoning of interest in diagnosis is called abductive inference (determination of the global most probable system description given the values of any partial subset of variables). In some cases, abductive inference can be performed with exact algorithms using distributed network computations but it is an NP-hard problem and complexity increases drastically with the presence of undirected cycles, number of discrete states per variable, and number of variables in the network. This paper describes an approximate method based on genetic algorithms to perform abductive inference in large, multiply connected networks for which complexity is a concern when using most exact methods and for which systematic search methods are not feasible. The theoretical adequacy of the method is discussed and preliminary experimental results are presented.

## 1 INTRODUCTION

Bayesian belief networks are graphs used to model uncertain systems by qualitatively and quantitatively encoding conditional dependence and independence among the system variables. Belief networks (BN) have a sound theoretical basis, are consistent with probability theory, and constitute a powerful tool in decision analysis and probabilistic reasoning in general. Recently developed methods for propagating probability information in the belief network structure have improved the ease with which probability data can be manipulated. These methods use distributed parallel computations in which probabilistic values are locally propagated between neighboring nodes in the belief network (Pearl 1988).

However, for large multiply connected networks, exact inference may not be feasible, rendering approximate algorithms an attractive alternative. Specifically, abductive inference with Bayesian belief networks is an NP-hard problem (Cooper 1990), and similarly, all the exact methods are highly sensitive to the connectedness of the networks (Horvitz 1990). Complexity increases with the number of variables in the system, the number of states per variable, and the number of undirected cycles in the network.

Different methods exist to find the most probable globally consistent explanation given the evidence. Pearl (1988) proposed an exact algorithm which can find the two best explanations for singly connected networks, but its growth is exponential for multiply connected networks. Shimony and Charniak (1990) obtain the maximum a-posteriori (MAP) assignment of values by using a best-first search on a modified belief network. The algorithm naturally extends to find next-best assignments, is linear in the size of polytrees but exponential in the general case. Peng and Reggia (1987) formalize causal and probabilistic associative knowledge in a two level network which associates disorders and manifestations. The structure is a special case of a belief network and calculations are computationally complex if multiple simultaneous disorders may occur.

This paper explores the use and performance of Genetic Algorithms (GAs) to find approximate near-optimal solutions in large and multiply connected belief networks. Section 2 summarizes the belief network framework and Section 3 describes the fundamentals of genetic algorithms. Section 4 discusses the adequacy of applying GAs to abductive inference in belief networks and describes in detail one genetic algorithm used. Section 5 describes four network examples and Section 6 presents experimental results on the performance of GALGO, an object-oriented implementation. A discussion of the results is presented in Section 7 along with future directions of research.

## 2 BELIEF NETWORKS

Belief networks consist of a set of propositional variables represented by nodes in a directed acyclic graph. Each variable can assume an arbitrary number of mutually exclusive and exhaustive values. Directed arcs between nodes represent the probabilistic relationships between nodes. The absence of a link between two variables indicates independence between them given that the values of their parents are known. In addition to the network topology, the prior probability of each state of a root node is required. For non-root nodes, the conditional probabilities of each possible value, given the states of its parent nodes or direct causes, are needed. Note that deterministic relations are a particular case which can be easily handled by having each conditional probability be either a 0 or a 1. Using the belief network framework, exact methods exist to perform abductive inference through the use of parallel calculations and message passing along the nodes in the network (Pearl 1988).

An important distinction must be made between singly and multiply connected networks. A directed acyclic graph is singly connected if there is at most one chain (or undirected path) between each pair of variables. Often the representation of physical systems results in multiply connected networks (with undirected cycles). Multiply connected networks require additional work to render them singly connected. Two well-known techniques are cutset conditioning (Pearl 1988) and clustering (Lauritzen 1988). Conditioning includes identifying the loops and selecting the minimal set of nodes whose instantiation breaks cycles. Clustering involves the aggregation of several nodes into a single node whose possible states are combinations of the states of the individual nodes. Methods for Gaussian continuous variables have also been proposed (Shachter 1989).

## 3   GENETIC ALGORITHMS

Approximate algorithms constitute a viable alternative when the size and topology of a problem render it intractable. The trade-off involves accepting a near-optimal solution in a feasible time. The method proposed in this paper to perform inference takes advantage of the BN framework to represent an uncertain system. The use of GAs to solve belief networks seems to be a simple yet powerful combination of a knowledge representation paradigm and an efficient inference engine. To the best knowledge of the authors, this approach has not been published before. To understand why GAs are particularly suited to perform inference in BNs, a review of the underlying concepts of these approximate algorithms is necessary.

Genetic algorithms are search procedures based on the mechanics of natural selection and natural genetics (Goldberg 1989). The methodology, architecture, and theoretical analysis were developed by Holland in 1975 (Holland 1992) for studying existing natural adaptive systems and designing artificial adaptive systems. The idea is that ..."adaptation can be usefully modeled as a form of search through a space of structural changes ... to improve its behavioral characteristics" (De Jong 1985).

The structural modification space is conventionally represented by strings of symbols chosen from some finite alphabet and the search within this space is accomplished by a procedure called a Genetic Algorithm (De Jong 1985).

GAs have several advantages over other methods. Conventional search methods are not robust, as discussed in (Goldberg 1989). GAs improve over the local scope of traditional methods (such as extrema determination in multidimensional spaces and gradient based search methods) by searching in parallel many subspaces in multidimensional spaces with complex topologies. Enumerative approaches are often not feasible or too slow for systems under time constraints. Random search algorithms simply lack the efficiency of GAs. According to Goldberg (1989) GAs differ from other methods in the following ways: (1) GAs work with a coding of the parameter set, not the parameters themselves, (2) GAs search from a population of points, not from a single point, (3) GAs use an objective function without any auxiliary knowledge, and (4) GAs use probabilistic transition rules, not deterministic rules.

The notion of *implicit parallelism* is a key strength of GAs and is responsible for the allocation of the search effort simultaneously to many hyperplanes (Grefenstette 1989). Hyperplanes can be thought of as subsets of points in the space which are consistent with a partial point specification. GAs process $O(N^3)$ hyperplanes in a population of size N. The practical implication of this result, along with a theoretical discussion of the underlying principles of GAs is presented in (Holland 1992; Goldberg 1989; Grefenstette 1989). The concept of a *schema* is used in the theoretical analysis of GAs by Holland (1992). A *schema* is a similarity template (a specification of alleles for a subset of the genes) describing a set of strings with similarities at certain positions. *Schemata* can be thought of as pattern matching devices. An important result (Holland 1992) is that highly fit, short-defining-length schemata, also called *compact building blocks*, are propagated through generations by giving exponentially increasing samples to the observed best individual.

## 4   THE ALGORITHM BEHIND GALGO

Conventional GAs use three genetic operators (1) selection, (2) crossover, and (3) mutation, which are introduced in this section. The proposed algorithm extends the conventional notion of GAs and can be described more accurately as *evolution programming* (Michalewicz 1992) since it uses non-binary alphabets, graphs instead of strings, and graph-operators.

### 4.1   Representation

A solution or individual is conventionally represented by a string of integers or chromosome which encodes the individual genotype. Each position or gene in the string corresponds to one variable in the belief network. Each gene can take a number of values or alleles from a finite discrete alphabet which may be different for each gene and corresponds to the number of discrete values that the variable can assume in the belief network. This first step is simple and is one of the factors which makes GAs so attractive for BNs. This simplicity notion is formalized as the *principle of minimal alphabets* (Goldberg 1989). The desirable representational feature of a low number of alleles per gene is naturally satisfied.

A non-conventional representation is used in this paper to represent a genotype. Since the evolution of individual solutions is based on the notion of successful compact blocks inherited through generations, it seems reasonable to construct these blocks in a form such that their elements are as closely related as possible. Adjacency in the graph corresponds to what can be called *semantic closeness*. It would be desirable to have neighboring nodes in the graph be neighbors in the genotype string. However, in the mapping from the graph to the string some nodes are necessarily separated. The structure which naturally satisfies this property is another graph. Therefore, in this algorithm, individuals in the population are graphs, and the elements of the compact blocks are almost always neighboring nodes.

GAs require the existence of a metric in the space of possible solutions. In this case there is a clearly defined metric, the absolute probability of each possible solution (or point in the search space, or system state in the BN space). Within the belief network framework performing this calculation is straightforward for the special case in which all the nodes have been instantiated (assigned a value) which is precisely the case that arises in this representation. The fitness metric corresponds to the individual phenotype and is a product with one factor for each node. Each factor is either a prior probability (for root nodes) or a conditional probability (for internal and leaf nodes). These probabilities are efficiently stored and retrieved using multidimensional arrays. To each genotype (set of variable-value assignments) corresponds a phenotype (fitness metric or probability).

### 4.2 Parameters

The GA algorithm requires the specification of several parameters. The main parameters quantify the population size, crossover rate and mutation frequency. The *total population* is the number of possible solutions and is usually a very large number. The *evolving population* is the number of individuals from which the final solution will evolve and is a small fraction, called *evolving fraction*, of the total population. The *breeding selectivity* is the fraction of the *evolving population* which constitutes the pool from which parents are chosen for breeding. The *breeding population* is the number of individuals which make up the parents pool. The *average lifetime* of individuals can be specified and is the average number of generations an individual stays in the evolving population set, before being displaced by better individuals. The *mutation frequency* is the fraction of the *evolving population* which suffers a mutation each generation. Mutations typically occur with a low frequency and consist of random changes introduced into the population to guarantee diversity and to prevent convergence to a local maximum. Finally the number of *generations* refers to the number of iterations in the algorithm. As suggested by (Grefenstette 1986) these parameters can be meta-optimized with another GA.

### 4.3 Selection

The initial population of individuals is created randomly and has a size much smaller than the total population. It is expected to randomly and uniformly sample all the search space. All individuals are guaranteed to have legal genotypes by assigning to each uninstantiated gene an allele (value) from the specific alphabet of each gene (variable). An arbitrarily specified fraction (1/average lifetime) of the evolving population is replaced in each generation by new individuals. Individuals with the lowest fitness are replaced. New individuals are created by combining parents, which are selected among the *best* found in the previous generation.

Selection of the *best* individuals can proceed according to different criteria. Three methods have been implemented. In the first, the probability of being selected as a parent (for each individual contained in the breeding population) is proportional to the phenotype of the individual. In the second method the probability of being selected is proportional to a monotonic function of the phenotype of the individual (i.e. $f(phenotype)=1/(\log^2(phenotype))$ ). This function can be used to control the sensitivity of the algorithm to fitness values. In the third method the probability for each individual is the same for all elements of the breeding population. This criterion reduces sensitivity to phenotype values. Sensitivity reduction can also be accomplished by using a function of the rank of each individual.

Immediately after each new individual is created, its phenotype or fitness is assessed and stored. In an iterative procedure, elements of each generation are sorted to choose which will be replaced and which will be used as parents for the next generation.

### 4.4 Crossover (reproduction)

New individuals are obtained by crossover of selected individuals of the previous generation. Two parents can create one or two children, being the latter the choice to avoid loosing potentially useful new individuals. The genotype of each new individual is made up by combining the genotypes of the parents. In traditional GAs two parents are copied into two children, two positions are randomly chosen in the new strings and the genes located between the two positions are interchanged. In this particular case, since individuals are represented as graphs, a *cluster*, which is a subset of the nodes in the BN, is interchanged. In the first step, each pair of children is created as a copy of the pair of parents. Second, a node is randomly chosen to be the center of the *cluster*. Determining the cluster elements is extremely simple since the only constraint is that all nodes located less than N links away from the center node be included. N is chosen to contain about half of the nodes in the cluster. In the third step the cluster is interchanged. By doing this the genotype of each resulting child is a combination of the genotypes of both parents. More elaborate cluster construction algorithms might improve performance by optimally identifying loosely connected components but these algorithms are expected to be computationally expensive.

### 4.5 Mutation

A mutation is a random change in one allele of the genotype of one individual. The mutation frequency is usually very low and its goal is to maintain diversity in the population to avoid premature convergence. The BN can be used for predictive reasoning or diagnostic abductive inference in which case any arbitrary subset of the variables may be instantiated (assigned a known value) during the inference process. Instantiated values are not changed by the mutation to guarantee that all individuals retain legal and meaningful genotypes.

### 4.6 Performance Evaluation

Useful measures of the performance of the algorithm include (1) the presence or absence of the optimal solution in the evolving population, (2) the distance between the best individual and the optimal one, (3) the accumulated probability mass in the best n solutions, (4) the fraction of offspring which improves over its ancestors, (5) on-line performance (the average fitness in the population), (6) off-line performance (fitness of the best individual)). Criteria (1) and (2) can be used for algorithm development purposes by comparing with the result of a systematic enumeration (only on small problems) of all possible solutions.

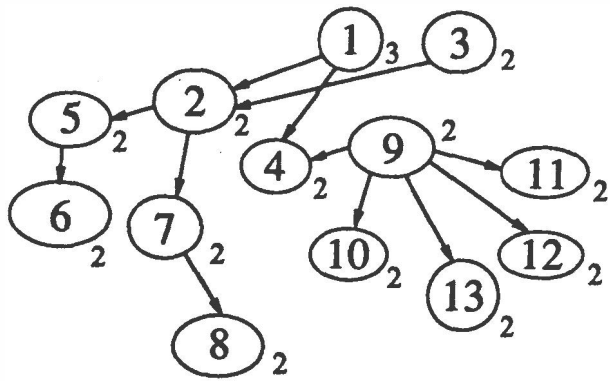

Figure 1: Topology of Belief Network 1

Table 1: Probability Distributions for Belief Network 1

| 1 | p(1) |   | 3 |   | p(3) |   | 9 |   | p(9) |
|---|------|---|---|---|------|---|---|---|------|
| 1 | 0.6  |   | 1 |   | 0.1  |   | 1 |   | 0.85 |
| 2 | 0.3  |   | 2 |   | 0.9  |   | 2 |   | 0.15 |
| 3 | 0.1  |   |   |   |      |   |   |   |      |

| 2 | 1 | 3 | p(2\|1,3) | 4 | 1 | 9 | p(4\|1,9) |
|---|---|---|-----------|---|---|---|-----------|
| 1 | 1 | 1 | 0.90 | 1 | 1 | 1 | 0.90 |
| 2 | 1 | 1 | 0.10 | 2 | 1 | 1 | 0.10 |
| 1 | 1 | 2 | 0.01 | 1 | 2 | 1 | 0.20 |
| 2 | 1 | 2 | 0.99 | 2 | 2 | 1 | 0.80 |
| 1 | 2 | 1 | 0.40 | 1 | 3 | 1 | 0.05 |
| 2 | 2 | 1 | 0.60 | 2 | 3 | 1 | 0.95 |
| 1 | 2 | 2 | 0.30 | 1 | 1 | 2 | 0.20 |
| 2 | 2 | 2 | 0.70 | 2 | 1 | 2 | 0.80 |
| 1 | 3 | 1 | 0.99 | 1 | 2 | 2 | 0.05 |
| 2 | 3 | 1 | 0.01 | 2 | 2 | 2 | 0.95 |
| 1 | 3 | 2 | 0.90 | 1 | 3 | 2 | 0.01 |
| 2 | 3 | 2 | 0.10 | 2 | 3 | 2 | 0.99 |

| 5 | 2 | p(5\|2) | 6 | 5 | p(6\|5) |
|---|---|---------|---|---|---------|
| 1 | 1 | 0.10 | 1 | 1 | 0.90 |
| 2 | 1 | 0.90 | 2 | 1 | 0.10 |
| 1 | 2 | 0.95 | 1 | 2 | 0.10 |
| 2 | 2 | 0.05 | 2 | 2 | 0.90 |

| 7 | 2 | p(7\|2) | 8 | 7 | p(8\|7) |
|---|---|---------|---|---|---------|
| 1 | 1 | 0.90 | 1 | 1 | 0.20 |
| 2 | 1 | 0.10 | 1 | 2 | 0.75 |
| 1 | 2 | 0.10 | 2 | 2 | 0.25 |
| 2 | 2 | 0.90 | 2 | 1 | 0.80 |

| 10 | 9 | p(10\|9) | 11 | 9 | p(11\|9) |
|----|---|----------|----|---|----------|
| 1 | 1 | 0.90 | 1 | 1 | 0.70 |
| 2 | 1 | 0.10 | 2 | 1 | 0.30 |
| 1 | 2 | 0.20 | 1 | 2 | 0.05 |
| 2 | 2 | 0.80 | 2 | 2 | 0.95 |

| 12 | 9 | p(12\|9) | 13 | 9 | p(13\|9) |
|----|---|----------|----|---|----------|
| 1 | 1 | 0.75 | 1 | 1 | 0.99 |
| 2 | 1 | 0.25 | 2 | 1 | 0.01 |
| 1 | 2 | 0.15 | 1 | 2 | 0.01 |
| 2 | 2 | 0.85 | 2 | 2 | 0.99 |

## 5 EXAMPLES

Four networks are used to illustrate the algorithm and to explore its performance. These examples have different sizes and degrees of connectivity. The first network, BN1, is an abstraction of a singly connected belief network model for a section of a chemical plant (Rojas-Guzmán 1992) whose topology is shown in Figure 1. This network has 13 nodes and its probability parameters were obtained from behavioral descriptions. The numbers inside the nodes are node identifiers and the small numbers outside each node indicate the number of discrete states of the node. The corresponding prior and conditional probabilities are included in Table 1. From a systematic enumeration of the 12,288 points which comprise the search space, the best (most probable) solution was found to be S=(1221111211111) with a probability of 0.098. The ordering of the genes (variables) in S corresponds to the numbering of the nodes (i.e. the value of node i is in position i). The best 100points (0.8% of the total 12,288 possible points) contain 62% of the probability mass.

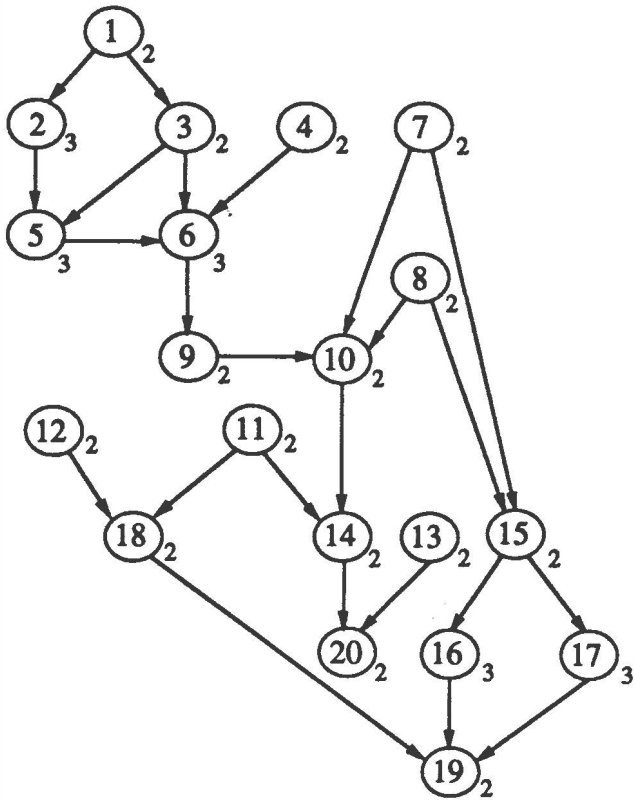

Figure 2: Topology of Belief Network 2

The second network, BN2 is shown in Figure 2. Both the topology and probability parameters of BN2 were generated randomly. BN2 has 20 nodes and represents a significantly larger search space with 7,962,624 possible states. This multiply connected network has 5 undirected cycles, 15 binary variables and 5 ternary variables, and its most probable state is S=(21212212211222122211) with p=5.98e-5. The third network, BN3 is shown in Figure 3. BN3 is a simplification of BN2 and has only one cycle. Its optimal solution is S=(21121212212222122121) with p=4.42e-5. The fourth network, BN4, has no links among variables but has the same search space size, and its optimal solution is S=(22211312121222223212) with p=3.15e-5.

## 6   EXPERIMENTAL RESULTS

Results from the proposed algorithm were compared with the solutions obtained by systematic exhaustive enumeration of all possible system states for BN1, BN2 and BN3. The best 50 solutions were stored in order in each run. Each network required approximately 70 hours on a 486 33MHz PC running a C++ implementation. The best solution for BN4 was simply calculated as the product of the largest prior probability of each node.

Results from 135 runs are summarized in Table 2. In all the runs, the average lifetime was set to 5 generations, which means that 20% of the individuals were replaced in each generation. Three parent selector criteria were used: a uniform probability distribution, a distribution proportional to the individual phenotype, and one proportional to a transformed value of the phenotype, where the transformation function is f(phenotype) = $1/(\log(phenotype))^2$. The mutation frequency was 0.025 for runs on BN1, and 0.075 for runs on BN2, BN3, and BN4. Each run required less than one minute.

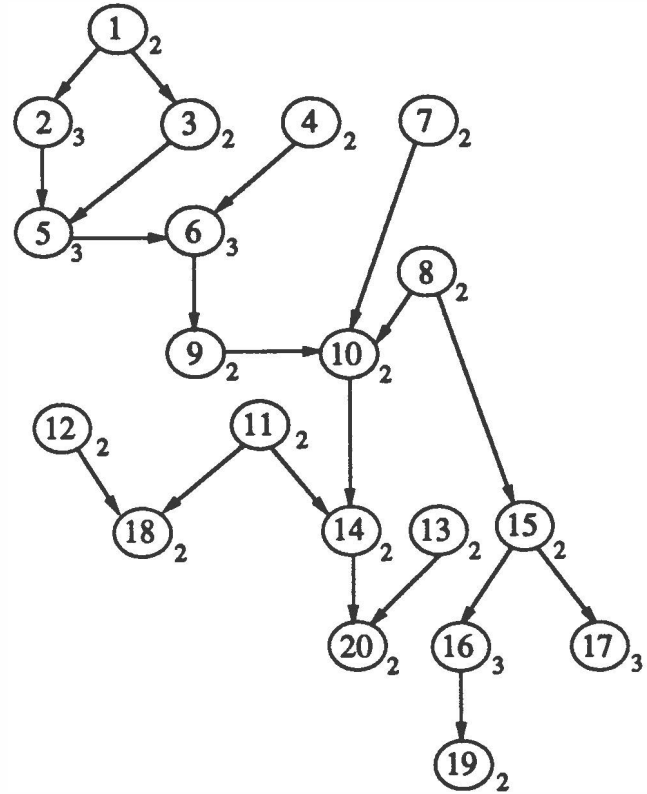

Figure 3: Topology for Belief Network 3

In Table 2, TOP N = X% means that in X% of the runs, a solution among the top N was obtained. Note that the set containing the top 50 solutions includes only 0.00063% of all the possible solutions for BN2. Rank refers to the average rank of the solutions to which the algorithm converged (rank=1 corresponds to the optimum). The standard deviations are also included. G indicates the average generation number at which the converged solution was first created. Gc corresponds to the average generation number at which convergence was reached. Figure 4 shows the evolution of the best phenotype. After a good solution is found, the population will take a few generations to converge as shown in Figure 5 where the evolution of the probability mass of the evolving population as a function of generations is plotted. The point at which the curve in Figure 5 becomes flat (generation 51) corresponds to convergence, the population is uniform and high frequency variations are due to mutations. In this specific run premature convergence was occurring around generation 40 on a genotype obtained in generation 26. As a result of a mutation the local optimum was avoided and the evolution converged to the global optimum in generation 51. EvalG indicates the number of individuals evaluated before G, and similarly, EvalGc is based on Gc. Note that Number of mutations = initial population + Generations * ((births/generation) + mutations/generation)). The size of the evolving population and the number of runs used are also indicated. Calculations to perform inference on networks with instantiated nodes are the same, except that mutations are not allowed on instantiated nodes. Note that complexity is a function of the number of non-instantiated nodes only.

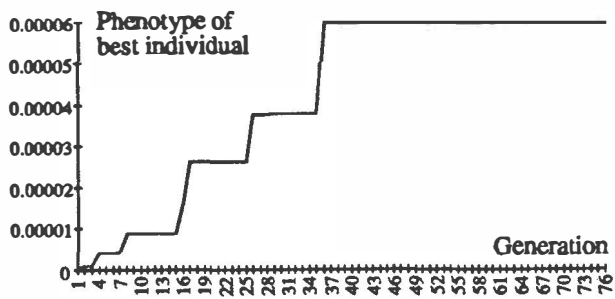

Figure 4: Phenotype of the best individual as a function of generations for BN2

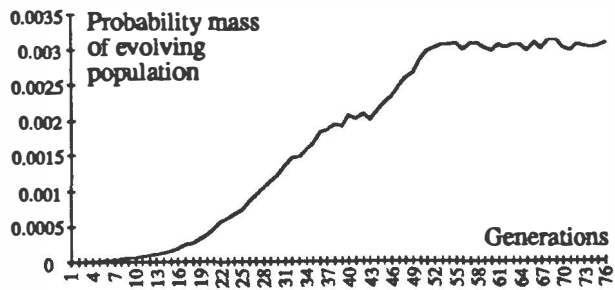

Figure 5: Probability Mass of Evolving Population as a function of generations for BN2

## 7 DISCUSSION

For large multiply connected networks, exact inference may not be feasible, rendering approximate algorithms an attractive alternative. Preliminary results have shown that Genetic Algorithms constitute a promising approach to perform inference in multiply connected complex systems. The proposed method yields sub-optimal (and often optimal) solutions in tractable times and avoids the strong sensitivity to the number of undirected loops in the network which makes exact methods not feasible for large models.

A random search to obtain the optimum with a probability of 0.20 (as in BN2 with the transformed phenotype selector) would have required evaluating 1.6 million points at each attempt, a significantly larger amount than the approximately 1000 evaluations required by the GA.

The complexity growth of the algorithm deserves careful attention and a large amount of experiments is required to obtain significant statistics. However, preliminary results comparing BN1 and BN2 are encouraging. BN1 is singly connected and represents a space of 12,288 states, whereas BN2 has 5 cycles, and has a significantly larger search space (7.96 million states). Nevertheless, convergence to solutions among the top N (for small N) required a similar number of point evaluations (around 1000).

Intuitively the sensitivity to the number of cycles in the GA approach would be small. Nevertheless performance is expected to be affected by the degree of connectivity in the network, but not particularly by the number of cycles. By comparing results from BN2 and BN4 using the transformed parent selector it is clear that in the extreme case of 0 arcs the problem is simpler and a greedy algorithm would be more efficient, as expected from considering connectivity and variable interactions.

There is a class of problems which is hard for GAs in general. From a practical and theoretical standpoint it is of interest to study the BN and GA combination proposed in this paper to determine whether hard problems are likely to arise and under which conditions this might happen. A problem is *deceptive* if certain hyperplanes guide the search toward some solution or genetic building block that is not globally competitive (Goldberg 1989). Whitley (1991) showed that the only problems which pose challenging optimization tasks are those that involve some degree of deception. According to Davidor (1991) three elements contribute to GA-hardness: (1) the structure of the solution space, (2) the representation of the solution space, and (3) the sampling error which results from finite and often small population sizes. By changing representations, GA-hardness may be diminished or avoided.

Davidor (1991) proposed the use of a statistic called *epistasis variance* to quantify the non-linearity in a representation. Epistasis (Klug 1986) refers to gene interactions. Some degree of interaction is necessary to guide the search in the space, but if interactions are too strong the problem will be hard. Zero epistasis would occur in a network without links. The best genotype could be found by a simple greedy algorithm following an approach similar to (Koutsoupias 1992) starting from a random position and changing genes, one at a time, to the allele which causes the largest improvement to the individual fitness. High epistasis would occur in a network with each node directly connected with all other nodes. A meaningful improvement in the fitness is expected to occur when all the nodes are simultaneously moved to the optimal. Fortunately, the structure which results in BNs has usually enough links to guide the search, and is very seldom fully connected. It is this local modularity (gene interactions are limited to immediate neighbors) which supports the notion of small compact blocks making a GA approach attractive over a greedy algorithm.

Table 2: Summary of Experimental Results

| | BN1 (0 cycles) Space size = 12,288 | BN2 (5 cycles) Space size = 7,962,624 | BN3 (1 cycle) Space size = 7,962,624 | BN4 (0 arcs) Space size = 7,962,624 |
|---|---|---|---|---|
| **UNIFORM** | TOP 1 = 95%<br>TOP 10 = 100%<br>TOP 50 = 100%<br><br>G < 50<br>Gc < 50<br>Eval G < 1310<br>Eval Gc < 1310<br>%Evaluated = 0.%<br>Ev.Population = 110<br>Total Runs = 20 | TOP 1 = 30%<br>TOP 10 = 60%<br>TOP 50 = 100%<br>Rank = 12.8, $\sigma$ = 16.3<br>G = 37.4, $\sigma$ = 9.4<br>Gc* = 49 (estimated)<br>Eval G = 785<br>Eval Gc = 1006<br>%Evaluated = 0.013%<br>Ev.Population = 75<br>Total Runs = 10 | | |
| **PROPORTIONAL** | TOP 1 = 30%<br>TOP 10 = 100%<br>TOP 50 = 100%<br>Rank = 1.05<br>G < 50<br>Gc < 50<br>Eval G < 1310<br>Eval Gc < 1310<br>%Evaluated < 10.7%<br>Ev.Population = 110<br>Total Runs = 10 | TOP 1 = 0%<br>TOP 10 = 45%<br>TOP 50 = 75%<br><br>G = 28.6, $\sigma$ = 12.5<br>Gc* = 41 (estimated)<br>Eval G = 618<br>Eval Gc* = 854<br>%Evaluated* = 0.011%<br>Ev.Population = 75<br>Total Runs = 20 | | |
| **TRANSFORMED** | | TOP 1 = 20%<br>TOP 10 = 88%<br>TOP 50 = 100%<br>Rank = 7.2, $\sigma$ = 10.8<br>G = 38.4, $\sigma$ = 15.0<br>Gc = 49.9, $\sigma$ = 11.6<br>Eval G = 805<br>Eval Gc = 1023<br>%Evaluated = 0.013%<br>Ev.Population = 75<br>Total Runs = 25 | TOP 1 = 8%<br>TOP 10 = 44%<br>TOP 50 = 56%<br><br>G = 45.6, $\sigma$ = 17.9<br>Gc = 58.2, $\sigma$ = 15.8<br>Eval G = 941<br>Eval Gc = 1180<br>%Evaluated = 0.015%<br>Ev.Population = 75<br>Total Runs = 25 | TOP 1 = 100%<br>TOP 10 = 100%<br>TOP 50 = 100%<br>Rank = 1.0, $\sigma$ = 0.0<br>G = 55.6, $\sigma$ = 16.9<br>Gc = 75.1, $\sigma$ = 17.9<br>Eval G = 1131<br>Eval Gc = 1501<br>%Evaluated = 0.019%<br>Ev.Population = 75<br>Total Runs = 25 |

Results indicate adequate convergence when parents are selected with a uniform probability and show premature convergence when the parent selection uses the proportional criteria due to the large differences in probabilities of solutions, especially at early stages in the evolution. A better parent selection which reduces sensitivity to phenotype values but still gives preference to individuals with higher phenotypes is based on the use of a transformed phenotype, as shown by comparing results from the three parent selection criteria on BN2.

The gene location within the string representation may be important for the existence (and consequent persistence) of building blocks. Allocating genes in a form such that neighbors in the belief network graph correspond to close genes in the chromosomal string has a theoretically appealing advantage but experiments are required to properly quantify the benefits of having *semantically close* compact blocks (by representing genotypes as graphs instead of strings).

Future work will explore two approaches for the optimization of the evolution parammeters of the GA. Solution accuracy and performance time can be combined to form a meta-fitness function. The second approach is based on a continuous revision of parameters as the evolution proceeds. Another area of reseach can exploit the efficient near-optimal global search of GAs together with some local search procedure to refine the solution once it is close to the optimum. According to the results found, location of the optimal solution by a small additional additional effort is possible. A local systematic search starting with each of the best n elements found can be performed by evaluating points within a specified small distance (measured as the sum of the differences between corresponding alleles).

Experiments are being conducted to characterize and compare the performance of the proposed algorithm on larger systems with different degrees of connectivity. Experiments to compare this approach with existing approximate algorithms will also be conducted.

This work was motivated by the requirements of real-time diagnostic reasoning tools for large, complex, and dynamic systems with strong non-linear interactions. Further research in this area is required to determine whether the proposed approach will prove practically useful to build decision support tools to diagnose and manage complex systems.